\documentclass{article}
\pdfoutput=1

\usepackage{arxiv}
\usepackage{graphicx}
\usepackage{amsmath}
\usepackage{multirow}
\usepackage{caption} 
\usepackage[utf8]{inputenc} 
\usepackage[T1]{fontenc}    
\usepackage{hyperref}       
\usepackage{url}            
\usepackage{booktabs}       
\usepackage{amsfonts}       
\usepackage{nicefrac}       
\usepackage{microtype}      
\usepackage{lipsum}

\title{Attention-Aware Age-Agnostic Visual Place Recognition}

\author{
  Ziqi Wang\thanks{These authors contributed equally in this work} \\
  Delft University of Technology\\
  \texttt{z.wang-8@tudelft.nl} \\
   \And
 Jiahui Li \footnotemark[1]\\
  Delft University of Technology\\
  \texttt{j.li-27@student.tudelft.nl} \\
  \And
 Seyran Khademi \\
  Delft University of Technology\\
  \texttt{s.khademi@tudelft.nl} \\
  \And
 Jan van Gemert \\
  Delft University of Technology\\
  \texttt{j.c.vangemert@tudelft.nl} \\
}

\begin{document}
\maketitle

\begin{abstract}
A cross-domain visual place recognition (VPR) task is proposed in this work, i.e., matching images of the same architectures depicted in different domains. VPR is commonly treated as an image retrieval task, where a query image from an unknown location is matched with relevant instances from geo-tagged gallery database. Different from conventional VPR settings where the query images and gallery images come from the same domain, we propose a more common but challenging setup where the query images are collected under a new unseen condition. The two domains involved in this work are contemporary street view images of Amsterdam from the \textit{Mapillary} dataset (source domain) and historical images of the same city from \textit{Beeldbank} dataset (target domain). We tailored
   an age-invariant feature learning CNN that can focus on domain invariant objects and learn to match images based on a weakly supervised ranking loss. We propose an attention aggregation module that is robust to domain discrepancy between the train and the test data. Further, a multi-kernel maximum mean discrepancy (MK-MMD) domain adaptation loss is adopted to improve the cross-domain ranking performance. Both attention and adaptation modules are unsupervised while the ranking loss uses weak supervision. Visual inspection shows that the attention module focuses on built forms while the dramatically changing environment are less weighed. Our proposed CNN achieves state of the art results ($99\%$ accuracy) on the single-domain VPR task and $20\%$ accuracy at its best on the cross-domain VPR task, revealing the difficulty of age-invariant VPR.
\end{abstract}

\keywords{Image Retrieval \and Domain Adaptation \and Attention Model}

\section{Introduction}
\begin{figure}[h]
    \centering
    \includegraphics[width=0.7\textwidth]{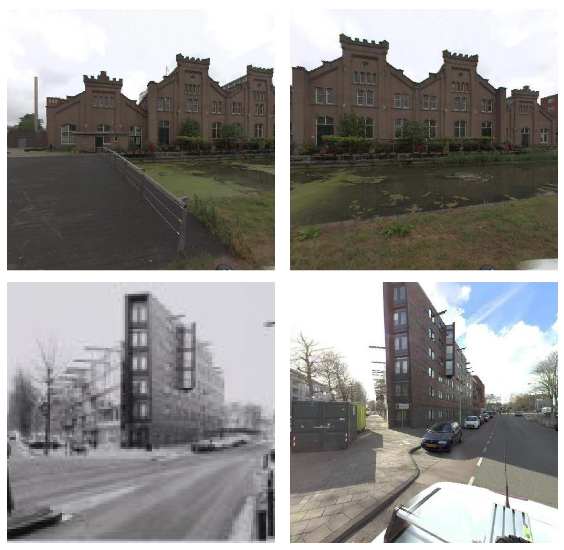}
    \caption{Correctly retrieved images with our proposed method. The top row illustrate the general place recognition in the same domain: both the query (left) and gallery image (right) are from the same dataset. The bottom row shows the cross-domain place recognition task where the query (left) is from the \textit{Beeldbank} dataset and the galley image (right) is from the \textit{Mapillary} dataset.}
    \label{fig:img1}
\end{figure}

Recently, there has been interest among the computer vision  researchers to solve the visual place recognition (VPR) task in the form of image retrieval \cite{Arandjelovic_2018,chen2017only,Iscen_2017, knopp_avoiding_2010,lopez2017appearance,torii_visual_2013,Zhu_2018}. In \cite{shi2019deep}, the discriminative visual cues learned for visual place classification task are investigated. Interestingly, CNN filters learn human-like discriminative visual cues to recognize a place, including built forms, signs or vegetation. Among these discriminative attributes, buildings are the most robust that remain, more or less, invariant during the changes in day and night lighting, different seasons and even years. However, CNNs are still influenced by irrelevant objects like roads and the sky. In this work, we introduce a CNN model with attention aggregation module to focus on domain invariant features, i.e. buildings, for the cross-domain VPR task. We will demonstrate that our work can be further combined with  multi-kernel Maximum Mean Discrepancy (MK-MMD) loss to obtain better domain adaptation results. The images from the two domains with a large time lag are depicted in Fig.\ref{fig:img1}, being historical images (queries) and current street view images (gallery) of Amsterdam. 
 
The VPR task is commonly formulated as content based image retrieval (CBIR), i.e., sorting the geo-tagged gallery images by their distances to the unknown query image. The query is then labeled based on its best matching image in the gallery. Deep image representation learning is currently state of the art for almost all CBIR settings.  Among the deep feature learning methods, distance learning CNNs are the most popular ones \cite{chopra2005learning,hoffer2015deep}. Nevertheless, supervised deep distance learning requires similar and dissimilar image pairs for training. In this work, image pair labels are not available and we only have access to geo-tagged images from the \textit{Mapillary} street view imagery and thus a weakly supervised deep feature learning is used, similar to the work of NetVLAD\cite{Arandjelovic_2018}.

Different from \cite{Arandjelovic_2018}, our queries are historical images which are not geo-tagged and exhibit a domain discrepancy between training data and test data. Age-agnostic place recognition that is addressed in this paper is a more challenging problem firstly due to the lack of image pair labels for training, secondly due to the domain shift between the gallery and query images caused by the change of scenery over a large time gap and thirdly due to the outliers in target domain. Different technologies of photography, equipment and processes used in the production of photos in the past also contribute to this domain shift. Fig\ref{fig:img1} shows the general and the age-agnostic place recognition task. 

\begin{figure}
    \centering
    \includegraphics[width=0.8\textwidth]{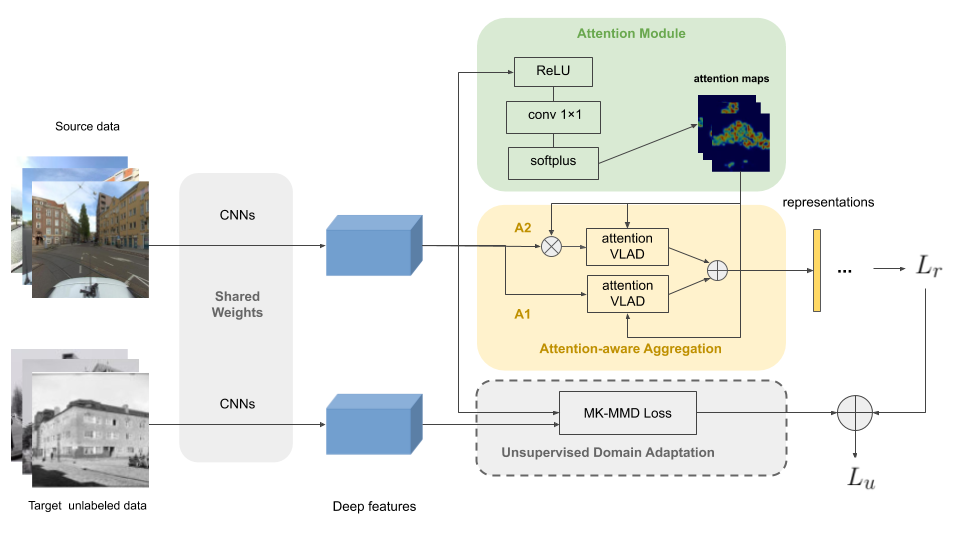}
    \caption{Our proposed CNN model include three modules: an attention module, am attention-aware VLAD module and a domain adaptation module. The attention-aware VLAD module uses the attention scores to weigh both the deep features and the global descriptors with two streams, $A_1$ and $A_2$, which are explained in Section~\ref{section: attention module}.}
    \label{fig:arch}
\end{figure}

We are inspired by \cite{nakka2018deep}, which introduces an attention module into NetVLAD for the classification task to address the unequal importance of local features in VLAD feature aggregation layer. In our work, a new attention aggregation technique is proposed to weigh both global VLAD descriptors and local descriptors. A domain adaptation loss based on MK-MMD is additionally introduced to achieve better cross-domain performance. Note that both the attention and the domain adaption modules are unsupervised and thus no labels are required. 

Our attention-aware architecture is depicted in  Fig.\ref{fig:arch} which consists of three modules and a shared convolutional neural network for feature extraction (AlexNet cropped before $conv5$). The attention module is a single convolutional layer followed by softplus activation function, transforming the feature map to a heatmap. This heatmap contains attention scores for the deep features. The VLAD module aggregates deep features in the attention-aware scheme by assigning attention scores to both local and global descriptors. The unsupervised domain adaptation module is additionally used to learn domain-invariant features. Our oblation studies show that both modules are important to reach state of the art results. Our speculation is that MK-MMD loss aligns the photo styles while attention module focuses on domain invariant contents.  

Our contributions are summarized as: 
\begin{itemize}
    \item  To the best of our knowledge, this is the first large scale (40k) image database for age-invariant visual place recognition task. We manually annotated 104 historical images and their corresponding matched current street view images only for evaluation purpose.
    \item A new attention aggregation scheme is proposed to combine both the local and global image descriptors (Section \ref{section: attention module}). 
    \item We combined the MK-MMD domain adaptation loss with the ranking loss to
    learn domain-invariant features for cross-domain VPR task. (Section \ref{section: MK-MMD}).
\end{itemize}

We tested our proposed model on conventional VPR task and our experiments show the state of the art results on Mapillary dataset compared to other competitors. Detailed results and ablation studies will be presented in Section \ref{section: results}. The comparison of single-domain and cross-domain results reveals the difficulty of age-agnostic place recognition task.

\section{Related Work}
The performance of VPR as an image retrieval problem depends on the ranking accuracy w.r.t. a similarity metric. The query location is suggested based on the top $M$ similar images (annotated with geo-tags). To extract good features for indexing, traditional works focus on hand-crafted features such as SIFT\cite{Lowe2004}) and SURF\cite{bay2006surf}. Some other efficient methods are based on the aggregation of local gradient-based descriptors like Fisher Vectors \cite{perronnin2010improving} and VLAD\cite{jegou2010aggregating}.  \cite{knopp_avoiding_2010} is a SURF based model which improves the performance by detecting and removing `confusing objects'. \cite{torii_visual_2013} uses SIFT to detect the repetitive patterns in the image which is representative for buildings. \cite{majdik_mav_2013} focuses on matching images that have large view point changes by generating artificial views of a scene for the training process.

Recent works suggest that a CNN trained on a large scale dataset as a feature extractor outperforms hand crafted features on various tasks \cite{chen2017deep,chen2017only, noh2017large,sarlin2019coarse}. In turn, \cite{Babenko_2014} shows that features in the early layers of a CNN trained for image classification can be effectively used as visual descriptors for image retrieval. LIFT \cite{Yi_2016} is a learning pipeline for feature extraction which introduces an end-to-end unified network for detection, orientation estimation, and feature description. \cite{tolias2015particular} proposes a global image representation by the regional maximum activation of convolutional layers (R-MAC) well-suited for place recognition. \cite{chen2017only} proposes novel CNN-based features designed for place recognition by detecting salient regions and extracting regional representations as descriptors. NetVLAD  \cite{Arandjelovic_2018} introduces a novel triplet ranking loss together with a VLAD aggregation layer that can learn powerful representations for the VPR task in an end-to end manner. A known disadvantage of NetVLAD lies in its global feature aggregation. \cite{gordo2016deep} proposes a region proposal network to learn which regions should be pooled to form the final global descriptor.
Similar to \cite{Arandjelovic_2018} , we use current geo-location tags for weakly supervised feature learning using triplet distance learning network. However, we do not have access to matched image pairs from the two domains for supervised training, i.e., matched historical and contemporary images. To address this domain mismatch between the test and train data, we need to promote domain-invariant feature learning. 

We tailor an attention aggregation model that can boost the cross-domain performance for our specific task, age-agnostic urban scene matching. Attention model is broadly used in natural language processing \cite{chorowski2015attention,vaswani2017attention} and computer vision tasks \cite{gu2018attentionaware,Long_2018,noh2017large, seymour2018semanticallyaware,song2017deep, Zhu_2018,xiao2015application}. \cite{kim2018attention} shows that attention model can also be adopted to benefit metric learning. \cite{noh2017large} proposes an attention mechanism to select key points for matching. Attention model is considered to be effective for domain adaptation as well \cite{Kang_2018,wang2019transferable}. Our attention model is implemented in an unsupervised way which means no ground truth score maps are available for training. The learning process of the attention module is guided by the image retrieval ranking loss. 

Given two different domains, unsupervised deep domain adaptation schemes \cite{long2015learning,tzeng2014deep} are mostly used to enhance the performance of CNNs on target domain by using labels only from the source domain. 

Among the vast amount of literature on deep domain adaption for classification tasks, the Maximum Mean Discrepancy (MMD) loss is introduced by \cite{borgwardt2006integrating} to minimize the domain discrepancy by projecting data into a kernel space. Later \cite{gretton2012optimal} proposed multi-kernel MMD (MK-MMD) which uses linear combination of multiple kernels. We adopt MK-MMD loss as an additional domain adaptation module for our attention aggregation model. Similarly, we feed untagged historical images of Amsterdam to the adaptation layer in an unsupervised manner.  


\section{Method}
Our proposed model consists of three modules for feature extraction, namely a weakly supervised image retrieval module with a triplet ranking loss (Section \ref{section: netvlad}), an attention aggregation module(Section \ref{section: attention module}) and an unsupervised domain adaptation module with MK-MMD loss (Section \ref{section: MK-MMD}). MK-MMD loss constrains the feature maps after the last convolution layer 
($conv\_5$). The final loss function for training, $L_{u}$, can be expressed as:

\begin{equation}
\label{eq:Lu}
    L_{u} =  L_r + \alpha M(\mathcal{D}_s,\mathcal{D}_t)
\end{equation}

where $M(\mathcal{D}_s,\mathcal{D}_t)$ is the MK-MMD loss term, $\mathcal{D}_s$ and $\mathcal{D}_t$ denote the source domain and target domain, $L_r$ is the triplet ranking loss used in NetVLAD \cite{Arandjelovic_2018}, $\alpha$ is the weight that trades off the image retrieval loss and the domain adaptation loss. 

\subsection{Image retrieval with weak supervision}\label{section: netvlad}
We use NetVLAD \cite{Arandjelovic_2018} as our baseline model which tackles the weakly supervised image retrieval task with a triplet ranking loss. NetVLAD considers the generated $H \times W \times D$ feature maps as a set of $N (H \times W) \times D$ local descriptors where N is the number of local descriptors and D is the dimension. Latter, a soft clustering is used to store the residual information contained in the descriptors to form $K \times D$ final descriptors denoted as $V$ where $K$ is the number of cluster centers. $V(j,k)$ can be expressed as:
\begin{equation}
\label{eq:equation1}
    V(j,k) = \sum^N_{i=1}a_k(x_i)(x_i(j)-c_k(j)), 
\end{equation}
where $j\in \{1,\dotsc, D\}$ is the $j$-th dimension of a descriptor $\{x_i\}$, $k \in \{1,\dotsc, K\}$ is the $k$-th cluster center, and $a_k(x_i)$ is the soft assignment of the descriptor $x_i$ to $k$-th cluster center $c_k$.
In Eq.\ref{eq:Lu}, A weakly supervised triplet ranking loss $L_r$ is used to govern the learning process of descriptors that ensures the Euclidean distance between the query image and the best potential positive images are smaller than the Euclidean distance between the query image and all the negative pairs (based on geo-tags).

\begin{equation}
\label{eq:Lr}
    L_r = \sum_j l~(\min_i d_\theta^2(q,p_i^q) + m - d_\theta^2(q,n_j^q)),
\end{equation}
where, $q$ denotes the query image and $p_i^q$ are potential positive images. $\min_id_\theta^2(q,p_i^q)$ denotes the best matching pair with shortest distance $d_\theta$. In turn, $n_j^q$ are all negative image pairs and $m$ is the distance margin to be maintained. The function $l$ is the hinge loss which penalizes the pairs that violate the margin. 


\subsection{Attention module}\label{section: attention module}
The triplet network for image retrieval task produces feature maps with the dimension of $ H \times W \times D$. 
The inserted attention module  consists  of  a $1 \times 1$  convolutional  layer with coefficients $\textbf{w}_a\in \mathbb{R}^{D\times1}$ and a softplus activation function. This convolutional layer will produce an attention score map $H_a$ with spatial size $ H \times W$, which could be interpreted as the weight $\{ w_i\}$ for each descriptor $\{x_i\}$. \cite{nakka2018deep} proposed an attention aware aggregation scheme $A_1$ as:

\begin{equation}
   V(j,k)_{A_1} = \sum^N_{i=1}w_ia_k(x_i)(x_i(j)-c_k(j)),
\label{eq:a1}
\end{equation}
where $w_i \in \textbf{w}_a$. Note that the VLAD module first assigns the local descriptors $\{x_i\}$ to $K$ cluster centers $\{c_k\}$, then computes the residuals of each descriptor $x_i-c_k$ to its cluster center and assigns the weight $a_k$ of descriptor $x_i$ to cluster $c_k$ proportional to their proximity.

In Eq.\ref{eq:a1}, the global descriptors (residuals) are weighed after clustering. However, the VLAD descriptor is very sensitive to cluster centers \cite{arandjelovic2013all} since it defines the origin of coordinates system to a cluster. Under this circumstance, we propose to weigh the local descriptors according to attention scores before performing clustering. The soft-assignment term $a_k$ is re-calculated based on the newly weighed descriptors. Our proposed aggregation scheme $A_2$ can be formulated as

\begin{equation}
\label{eq:a2}
   V(j,k)_{A_2} = \sum^N_{i=1}w_ia_k(x_iw_i)(x_i(j)w_i-c_k(j)).
\end{equation}{}
The difference between $A_1$ and $A_2$ is that $A_1$ assigns the attention scores after clustering the descriptors to multiple centers so the attention scores are only used to weigh the residuals but $A_2$ first uses the attention scores to filter out uninteresting regions in the individual local descriptors and then performs the same step as $A_1$. Considering that the reweighing of individual descriptors may remove information that are useful for global descriptor generation, we aggregate the two attention schemes linearly:
\begin{equation}
\label{eq: our a}
    V(j,k)_{our}= V(j,k)_{A_1}+V(j,k)_{A_2}. 
\end{equation}{}

\subsection{Domain adaptation module}\label{section: MK-MMD}
We use MK-MMD loss  \cite{gretton2012optimal}
with five Gaussian kernels of different bandwidths for unsupervised domain adaptation. The loss
minimizes the distance between the expectation of the kernel mappings $\phi(.)$ of the descriptors in the source domain $x_i^s$ and the target domain $x_i^t$.   
\begin{equation}
\label{eq:MK-MMD}
    M(\mathcal{D}_s,\mathcal{D}_t) = \sum_i^N||\mathbb{E}(\phi(x_i^s)) - \mathop{\mathbb{E}}(\phi(x_i^t))||_2.
\end{equation}

The MK-MMD loss guides the CNN to learn a latent space where the two domains are not distinctive, i.e., the gap between the statistical means of these two domains are closed in the reproducing kernel Hilbert space (RKHS). 

\section{Experiment}
\subsection{Dataset}
We construct a cross-domain dataset with two sources of data to evaluate our proposed method, namely the street view panorama images of  Amsterdam city from the \textit{Mapillary} dataset\cite{mapilllary} and the \textit{Beeldbank} dataset\cite{beeldbank} containing historical images from Amsterdam city archives.

 \begin{table}[h]
\centering
\begin{tabular}{c|c|cc}
\toprule
\multicolumn{2}{c|}{Dataset}    & Gallery & Query \\ \midrule
\multirow{2}{*}{Source} & \textit{Mapillary40k}-train & 20,884   & 2,320 \\ 
                        & single-domain-test  & 18,980 (\textbf{M})   & 2,108 (\textbf{M}) \\ \hline
\multirow{2}{*}{Target} & \textit{Beeldbank}-train    & 29,726   & -     \\ 
                        & cross-domain-test     & 2,469 (\textbf{M})    & 104 (\textbf{B})   \\ 
\bottomrule
\end{tabular}

\caption{\textit{Mapillary40k$\xrightarrow{}$Beeldbank dataset}, the source domain is \textit{Mapillary40k} and the target domain is \textit{Beeldbank} denoted by \textbf{M} and \textbf{B}, respectively. \textit{Beeldbank-train} is only used for unsupervised domain adaptation. cross-domain VPR requires matching query images from \textit{Beeldbank} to gallery images from \textit{Mapillary40k}.}
\label{table:cross-domain vpr dataset}
\end{table}

\textbf{\textit{Mapillary40k}} is a subset of  \textit{Mapillary250k} dataset collected from the source domain. The source domain contains panoramic images with high resolution collected from the \textit{Mapillary}, Amsterdam area. Each image is annotated with a geotag. The cylindrical panorama is converted to 6 cubmaps (all share the same geotag): `top', `down', `left', `right', `front' and `back' textures with 512 $\times$ 512 resolution. The `top' and `down' textures are discarded since they usually contain sky and the vehicle that carries the camera. 40k gallery images and 4k query images are collected in total which are then divided into two roughly equal parts for training and testing when tested for single-domain VPR task, each containing around 20k gallery images and 2k queries. The two sub-datasets are geographically disjoint. 

\textbf{\textit{Beeldbank}}, the target domain, contains historical images of Amsterdam with low resolution and random size (height and weight are around 100 pixels). This dataset not only depicts Amsterdam street view in the past but also contains outliers including people, sketches and indoor scenes.

\textbf{\textit{Mapillary40k - Beeldbank}} dataset is introduced in this work for the cross-domain VPR task (Table \ref{table:cross-domain vpr dataset}). The cross-domain test set contains 104 labeled queries from the target domain and 2,469 gallery images from the source domain. In the cross-domain test set, each target query has around 10 corresponding matched images in source domain. 30k unlabeled \textit{Beeldbank} images are used during training for domain adaptation.

\subsection{Single-domain and Cross-domain VPR tasks} 
\textbf{Single-domain VPR task ($S\xrightarrow{}S$)} In the single-domain VPR setup, we train the network with only weakly labeled source domain images. The network is tested on the test set of the same domain. \textit{Mapillary40k} is used for this single-domain VPR experiment as shown in Tab.\ref{table:cross-domain vpr dataset}. This is the common setting for VPR task as there is no domain mismatch between train and test data. We use single-domain VPR as a pilot experiment to evaluate the performance of the proposed model on conventional VPR task. 

\textbf{Cross-domain VPR task ($S\xrightarrow{}T$)}  The domain discrepancy between train and test data makes the cross-domain VPR task more challenging. This is the core of our experiments in this work which aims at labeling the images from beeldbank dataset with correct geo-location. We train the MK-MMD layer with weakly labeled source data and unlabeled target data for the cross-domain VPR task.  Labeled data with matching pairs from beeldbank and Mapilary dataset is only used for  evaluation of the model. We use queries from the target domain to retrieve relevant gallery image(s) collected from the source domain. 

We made the hypothesis that our attention module itself can improve the cross-domain VPR task to some extent without the MK-MMD loss compared to vanilla NetVLAD. An experiment was carried out to examine the function of the attention module later in Section \ref{section: attention module}. Further ablation study of the attention module and the MK-MMD loss will be presented in section \ref{section: results- attention module} and \ref{section: results-DA}. 

\textbf{Baseline work}  We compare our attention-aware framework with `off-the-shelf' CNNs for both single-domain VPR and cross-domain VPR tasks. The baseline work used AlexNet pretrained on ImageNet cropped before $conv5$ as feature extractor. Features are then sub-sampled by either max pooling ($f_{max}$), average pooling ($f_{avg}$), vanilla VLAD pooling without attention($f_{VLAD}$) and VLAD with attention-aware $A_1$ method ($f_{A1-VLAD}$) \cite{nakka2018deep}.

\subsection{Evaluation metrics}
We follow the standard place recognition evaluation metric in \cite{Arandjelovic_2018} where the query image is considered as correctly matched if at least one of the retrieved top $N$ images is located within 25 meters away from the ground truth query location. The Recall@$N$ evaluates the percentage of correctly localized queries at different $N$ matching levels. For cross-domain place recognition, since the \textit{Beeldbank} dataset contains labeled positive pairs, the Recall@$N$ will be directly calculated using these labels. 

\begin{figure*}[h!]
\centering
  \begin{tabular}{@{}cccccccc@{}}
    \rotatebox{90}{\qquad \qquad Query}&
    \includegraphics[width=.2\textwidth]{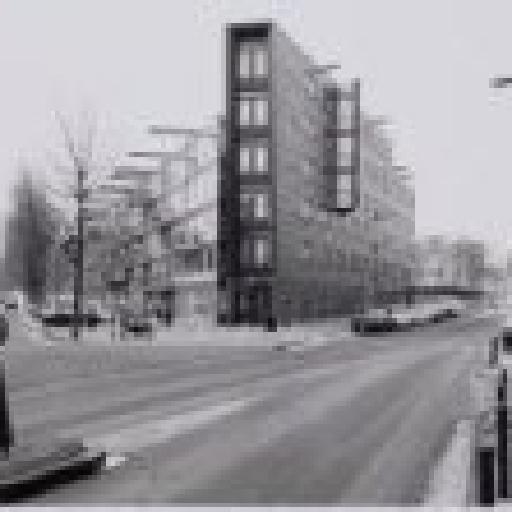} &
    \includegraphics[width=.2\textwidth]{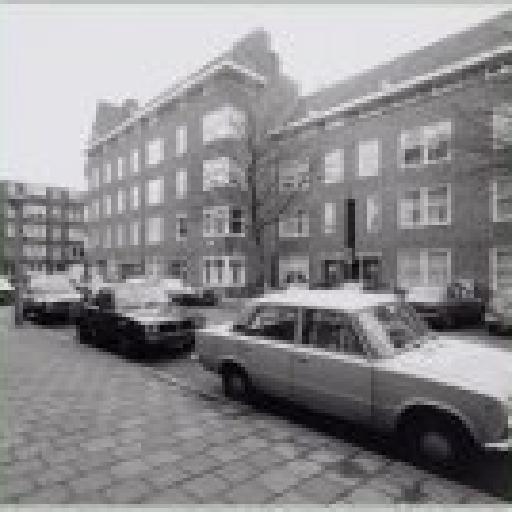} &
    \includegraphics[width=.2\textwidth]{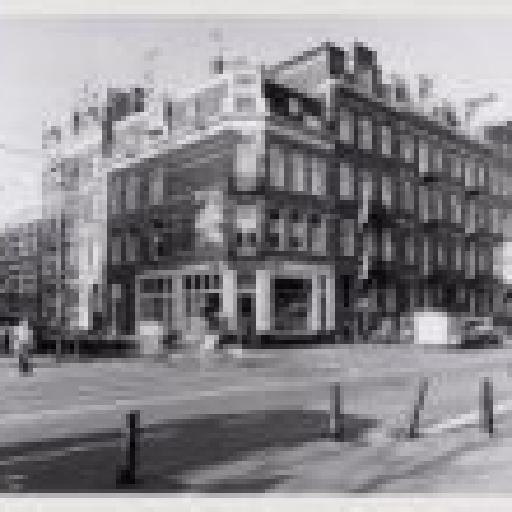} &
    \includegraphics[width=.2\textwidth]{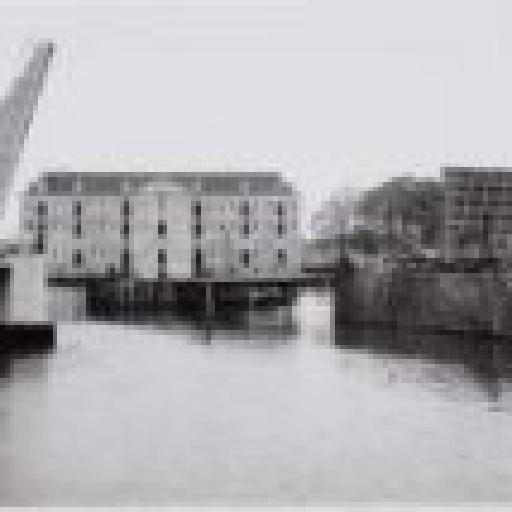} &\\
    \rotatebox{90}{\quad \quad Top1 Image}&
    \includegraphics[width=.2\textwidth]{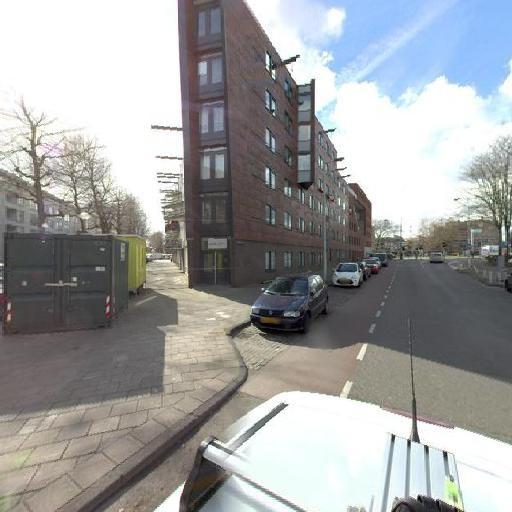} &
    \includegraphics[width=.2\textwidth]{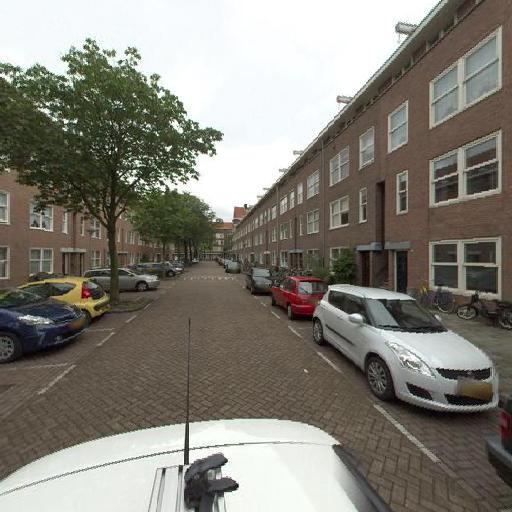} &
    \includegraphics[width=.2\textwidth]{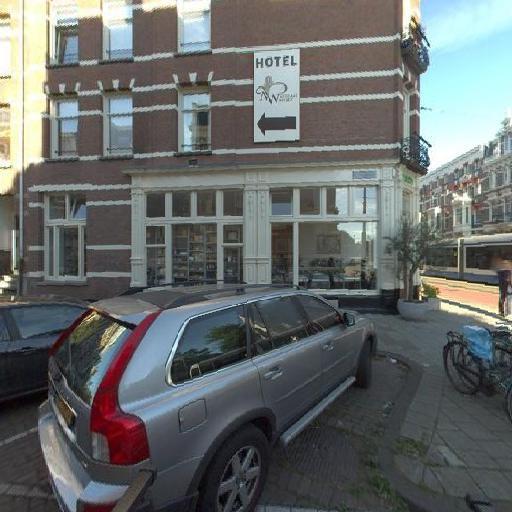} &
    \includegraphics[width=.2\textwidth]{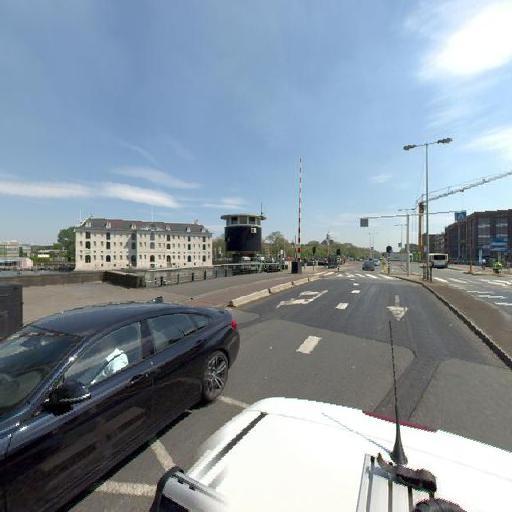} &\\
    &a. &b. &c. &d.
  \end{tabular}
  \caption{Correctly retrieved  top1 image from our framework trained with unsupervised domain adaptation, (top) queries are from the \textit{Beeldbank} dataset, (bottom) retrieved images are from the \textit{Mapillary} dataset. Our model can retrieve images not only depicting a similarly scene (a.), but also images from a different perspective (c.) and images captured further away from the query (b., d.). (b.) is correctly retrieved by matching the features of the building like the window and the unique shape of the door on the right side. }
  \label{fig:cross}
\end{figure*}

\begin{figure*}[h!]
\centering
  \begin{tabular}{@{}cccccccc@{}}
  & Target & Source\\
     \rotatebox{90}{\quad \quad Input}&
    \includegraphics[width=.15\textwidth]{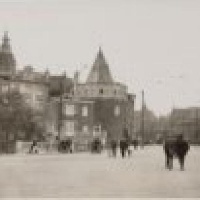}
    \includegraphics[width=.15\textwidth]{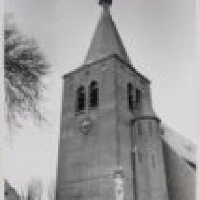}
    \includegraphics[width =.15\textwidth]{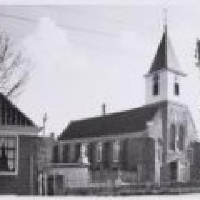}
    &
    \includegraphics[width=.15\textwidth]{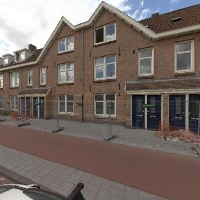} 
    \includegraphics[width=.15\textwidth]{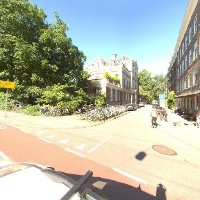} 
    \includegraphics[width=.15\textwidth]{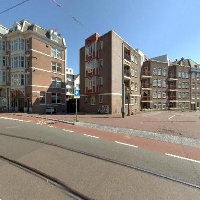} 
    \\
    \rotatebox{90}{ \qquad \quad $A1$} &
    \includegraphics[width=.15\textwidth]{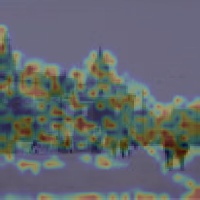}
    \includegraphics[width=.15\textwidth]{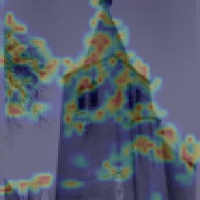} 
    \includegraphics[width =.15\textwidth]{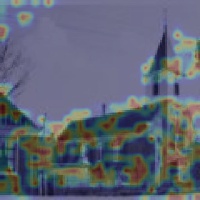}
    &
    \includegraphics[width=.15\textwidth]{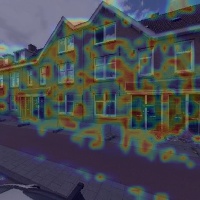} 
    \includegraphics[width=.15\textwidth]{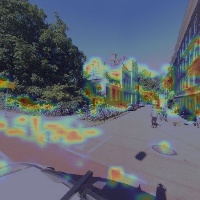} 
    \includegraphics[width=.15\textwidth]{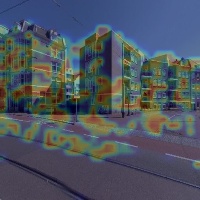} 
  \\
    \centering {\rotatebox{90}{\qquad \quad $Our$}}&
    \includegraphics[width=.15\textwidth]{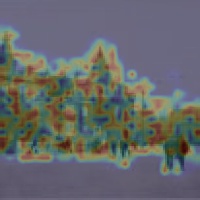}     \includegraphics[width=.15\textwidth]{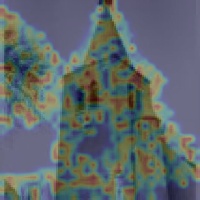} 
    \includegraphics[width =.15\textwidth]{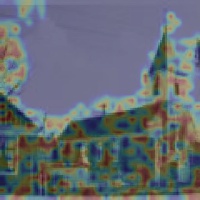}
   &
    \includegraphics[width=.15\textwidth]{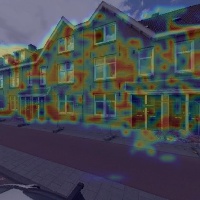} 
    \includegraphics[width=.15\textwidth]{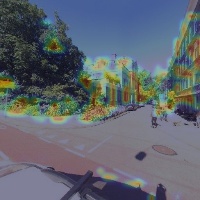} 
    \includegraphics[width=.15\textwidth]{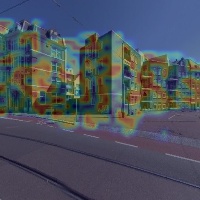} 
  \end{tabular}
  \caption{Visualization of attention score maps for source and target images. The top row shows the input images. The middle row is the heatmaps obtained by using \cite{nakka2018deep}, defined in Eq.\ref{eq:a1}. The bottom row presents the results from our proposed method defined in Eq.\ref{eq: our a}. It shows that our proposed attention module can generate accurate attention score maps with higher density on domain invariant objects for both source images and target images.}
  \label{fig:heatmap}
\end{figure*}

\subsection{Implementation Details}
The attention module starts with a ReLU activation, followed by a $1\times 1$ convolutional layer and softplus activation to produce attention scores. In the VLAD layer, the number of cluster centers used is $K = 64$. \textit{Mapillary} images were cropped with a random proportion to the original size between (0.3$~$1.0) for data augmentation before training.

We froze the layers before $conv4$ and fine-tuned the weights of all the other layers afterwards with the optimizer ADAM. We used the following hyperparameters: learning rate $lr=1e\text{-}5$, batch size = 2 tuples (each tuple contains 24 images, including query, positive and negative pairs), epochs = 25. The hard negatives mining uses the same technique as NetVLAD\cite{Arandjelovic_2018}: it first caches all the training queries and gallery images for a time and then randomly selects 1000 negatives (image away from 25 meters). It keeps the top 10 hardest negatives from the cached gallery image features. The cache is updated every 1000 training queries.

We center cropped and reshaped all target \textit{Beeldbank} images to $512\times512$ pixels in the cross-domain VPR experiment. The MK-MMD loss is calculated after $conv5$. The weight $\alpha$ in Eq.\ref{eq:Lu} is 0.99. The margin $m$ in Eq.\ref{eq:Lr} is set as 0.1.

\subsection{Results}\label{section: results}
This section presents the results of the experiments with a detailed ablation study for the attention module (Section \ref{section: results- attention module}) and the domain adaptation module (Section \ref{section: results-DA}) separately on both single-domain and cross-domain VPR tasks. Visual inspection of retrieval results and attention heatmaps are shown in Fig.\ref{fig:cross} and Fig.\ref{fig:heatmap}.

\subsubsection{Attention module}\label{section: results- attention module}
To evaluate the performance of our attention aggregation module on both single and cross-domain VPR tasks, we first trained the model on the source domain (\textit{Mapillary40k}) and directly tested it on the source test set and the target test set without MK-MMD loss. Tab.\ref{tab:comparison3} shows the retrieval results where our attention aggregation method consistently outperforms the model without attention on both $S \xrightarrow{} T$ and $S \xrightarrow{} S$ tasks. A possible explanation could be that the VLAD descriptors are easily affected by the irrelevant objects. By not focusing on representative details that describe unique features of each building, it may retrieve an image that has a similar road or sky etc.  
\begin{table*}[h]
\centering
\begin{tabular}{c|cccc|cccc}
\toprule
& \multicolumn{4}{c|}{$S\xrightarrow{} T$} & \multicolumn{4}{c}{$S\xrightarrow{} S$}\\
 &     R@1  & R@5    &R@10   &R@20    &  R@1  & R@5 &R@10   &R@20   \\
\hline
$f_{max}$$^+$  &0.0096 & \textbf{0.0577}& 0.0769 &0.1058  &  0.6347 & 0.8226&  0.8800& 0.9203 \\ 
$f_{avg}$$^+$  & 0.0000  & 0.0096 &0.0481 & 0.0769 & 0.7884& 0.9284& 0.9535& 0.9730\\
$f_{max}$    &  0.0000 & 0.0000 & 0.0577 &0.1250 &  0.7410& 0.9108& 0.9431& 0.9639 \\
$f_{avg}$    &  0.0096 &0.0192 &0.0481 & 0.0577 & 0.7984& 0.9269&0.9564& 0.9725   \\
$f_{VLAD}$ & 0.0096& 0.0192& 0.0192& 0.0577 & 0.8843 & 0.9687& 0.9782& 0.9853\\
$f_{A1\text{-}VLAD}$  & 0.0096& 0.0481& 0.1058& 0.1538 & 0.8819& 0.9649& 0.9801& 0.9877\\
$f_{our\text{-}VLAD}$ & \textbf{0.0192} &  \textbf{0.0577} &  \textbf{0.1154}& \textbf{0.2019} & \textbf{0.9132}& \textbf{0.9753}& \textbf{0.9815}& \textbf{0.9900} \\
\bottomrule
\end{tabular}
\caption{The $^+$ denotes that the `off-the shelf' model is pretrained on ImageNet\cite{deng2009imagenet} for classification task. The others are trained on \textit{Mapillary40k} for place recognition from scratch, and directly tested on the cross-domain dataset.}
\label{tab:comparison3}
\end{table*}

To inspect whether our attention module can produce reasonable attention scores for each descriptor, we visualize the attention maps of different attention-aware schemes in Fig.\ref{fig:heatmap}. Our attention aggregation method generates heatmaps with higher densities on representative features and better robustness against irrelevant objects. Most attention is assigned to the architectures and less attention is assigned to non representative regions such as road and sky as expected. Note that in Tab.\ref{tab:comparison3}, the performance of $f_{A1\text{-}VLAD}$ is worse than $f_{VLAD}$ and $f_{our\text{-}VLAD}$ achieves the best results. We conclude that an insufficient attention map will deteriorate the performance.

\subsubsection{Domain adaptation module} \label{section: results-DA}
\begin{figure}[h]
    \centering
    \includegraphics[width=0.7\textwidth]{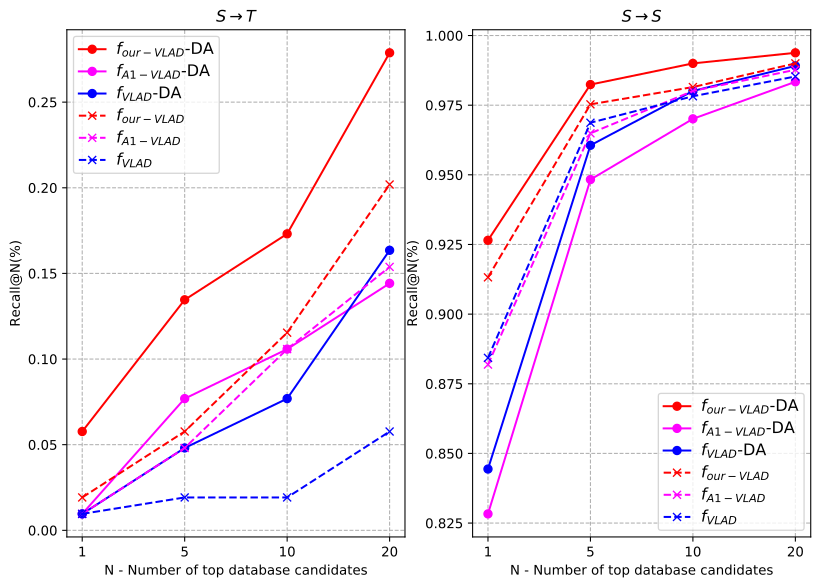}
    \caption{Comparison of the models trained with or without the MK-MMD loss on both single-domain ($S \xrightarrow{} S$) and cross-domain ($S \xrightarrow{} T$) tasks. DA denotes that the MK-MMD loss is added during training}
    \label{fig:da}
\end{figure}

The additional domain adaptation loss (MK-MMD) is added to our model and all baseline works in this section. The MK-MMD loss is adopted to further minimize the domain discrepancy in this experiment. We applied it on the vanilla NetVLAD ($f_{VLAD}\text{-}DA$), $A_1$ attention model ($f_{A1\text{-}VLAD\text{-}DA}$) and our attention aggregation model ($f_{our-VLAD}\text{-}DA$). The performance of different models with and without MK-MMD loss are examined on both source and target test test. The results are visualized at different recall rates in Fig.\ref{fig:da}.

When trained with the MK-MMD loss for the $S\xrightarrow{}T$ cross-domain VPR task, both $f_{VLAD}\text{-}DA$ and $f_{our\text{-}VLAD}\text{-}DA$ benefit from domain adaptation, while no significant improvement of $f_{A1\text{-}VLAD\text{-}DA}$ is observed. Detailed results are presented in Tab.\ref{tab:DA}. 

\begin{table*}[h]
\centering
\begin{tabular}{c|cccc|cccc}
\toprule
& \multicolumn{4}{c|}{$\text{with DA}$} 
& \multicolumn{4}{c}{$\text{without DA}$}\\
 &     R@1  & R@5    &R@10   &R@20    &  R@1  & R@5 &R@10   &R@20   \\
\hline
$f_{VLAD}$  &0.0096&0.0481& 0.0769& 0.1635  &  0.0096& 0.0192& 0.0192& 0.0577 \\ 
$f_{A1\text{-}VLAD}$  & 0.0096&0.0769&0.1058&0.1442 & 0.0096& 0.0481& 0.1058& 0.1538\\
$f_{our\text{-}VLAD}$ & \textbf{0.0577} &  \textbf{0.1346} &  \textbf{0.1731}& \textbf{0.2788} & \textbf{0.0192} &  \textbf{0.0577} &  \textbf{0.1154}& \textbf{0.2019} \\
\bottomrule
\end{tabular}
\caption{Comparison of different models' performance on the cross-domain $S \xrightarrow{} T$ VPR task under two conditions: with or without domain adaptation using the MK-MMD loss. DA stands for domain adaptation using MK-MMD loss.}
\label{tab:DA}
\end{table*}

In addition, we also examined the performance of the model trained for the cross-domain VPR task on the source domain due to the reason that extra data from the target doamin may also help with retrieval in source domain if the model is robust to the outliers in the \textit{Beeldbank} dataset. $f_{VLAD}\text{-}DA$ does not show much power in the original source domain compared to $f_{VLAD}$. The retrieval accuracy of $f_{A1\text{-}VLAD\text{-}DA}$ in the source domain decreases after domain adaptation. Our proposed model gets better retrieval result even on the source domain as shown in Fig.\ref{fig:da}. This experiment proves that the domain specific features and outliers are reduced while more domain invariant features are captured by our proposed attention aggregation model which further facilitates the domain adaptation procedure. 

Overall, we show that our attention aggregation model can achieve more accurate retrieval results on both single-domain $S \xrightarrow{} S$ and cross-domain $S \xrightarrow{} T$ VPR tasks even without domain adaptation and it can further facilitate unsupervised domain adaptation to achieve better performance on both source and target test sets.

\section{Discussion}
Usually we assume that the training data and test data are sampled from an identical distribution which is violated in our cross-domain setting. We designed an attention-aware adaptive network to tackle the existing distribution shift. The results indicate that both the attention and adaptation modules contribute to the accurate retrieval of visual information. We speculate that the attention module mainly helps with focusing on domain invariant objects and the domain adaptation module aligns the depiction styles between the two different domains. Our dual experiments on both conventional and cross-domain VPR tasks admit the difficulty of learning age-invariant features when there is no cross-domain pairing labels available for directly training CNNs.   

Besides the large domain shift, our \textit{Beeldbank} target dataset contains various classes of images like people, indoor scenes, sketches and ground plans of buildings. These outliers are not contained in source dataset \textit{Mapillary40k} rendering the task more difficult. Domain adaptation with more classes or outliers in the target domain compared to the source domain can be considered as open-set domain adaptation problem \cite{Baktashmotlagh2019OPENSETDA,Bendale_2016,oza2019deep,Saito_2018}. Some other works refer to this as outlier detection problem \cite{yamaguchi2019adaflow,luo2018discriminative}. We speculate that the attention module can filter out the outliers by weighing them less with the heatmaps. 

\section{Conclusion}
We proposed a specially-designed CNN for automatic annotation of historical images with their location. This is helpful specifically for museum curators and historians to retrieve the location information of a historical urban scene or architecture.  This task is more challenging than single-domain (conventional) location retrieval due to the domain discrepancy caused by the large time lag between depicted scenes. A cross-domain dataset is collected accordingly with \textit{Mapillary40k} used as source domain and \textit{Beedldbank}, as target domain. To tackle this challenge, an attention aggregation module with a domain adaptation layer is designed, the performance of which is demonstrated by detailed experiments and ablation studies. Our attention aggregation model achieves state of the art results on both single and cross-domain VPR tasks by focusing more on domain invariant objects. It can be further combined with an extra domain adaptation module using the MK-MMD loss to achieve higher retrieval accuracy not only on the target domain but also on the source domain.  Moreover, we believe our methods can achieve promising results on open-set domain adaptation tasks where unseen classes or outliers are not involved during training.

\section*{Acknowledgements}
We would like to thank Dr. Ronald Siebes for collecting the Mapillary dataset and other members of ArchiMedial project \url{http://archimedial.eu/} which is funded by Volkswagen Foundation in Germany.   
This work is also partly funded by the Netherlands Organization for Scientific Research (NWO) under the research program C2D–Horizontal Data Science for Evolving Content with project name DACCOMPLI and project number 628.011.002. 

\bibliographystyle{unsrt}  
\bibliography{references}  

\begin{thebibliography}{10}

\bibitem{Arandjelovic_2018}
Relja Arandjelovic, Petr Gronat, Akihiko Torii, Tomas Pajdla, and Josef Sivic.
\newblock Netvlad: Cnn architecture for weakly supervised place recognition.
\newblock {\em IEEE Transactions on Pattern Analysis and Machine Intelligence},
  40(6):1437–1451, Jun 2018.

\bibitem{chen2017only}
Zetao Chen, Fabiola Maffra, Inkyu Sa, and Margarita Chli.
\newblock Only look once, mining distinctive landmarks from convnet for visual
  place recognition.
\newblock In {\em 2017 IEEE/RSJ International Conference on Intelligent Robots
  and Systems (IROS)}, pages 9--16. IEEE, 2017.

\bibitem{Iscen_2017}
Ahmet Iscen, Giorgos Tolias, Yannis Avrithis, Teddy Furon, and Ondřej Chum.
\newblock Panorama to panorama matching for location recognition.
\newblock {\em Proceedings of the 2017 ACM on International Conference on
  Multimedia Retrieval - ICMR ’17}, 2017.

\bibitem{knopp_avoiding_2010}
Jan Knopp, Josef Sivic, and Tomas Pajdla.
\newblock Avoiding confusing features in place recognition.
\newblock In {\em European Conference on Computer Vision}, pages 748--761.
  Springer.

\bibitem{lopez2017appearance}
Manuel Lopez-Antequera, Ruben Gomez-Ojeda, Nicolai Petkov, and Javier
  Gonzalez-Jimenez.
\newblock Appearance-invariant place recognition by discriminatively training a
  convolutional neural network.
\newblock {\em Pattern Recognition Letters}, 92:89--95, 2017.

\bibitem{torii_visual_2013}
Akihiko Torii, Josef Sivic, Tomas Pajdla, and Masatoshi Okutomi.
\newblock Visual place recognition with repetitive structures.
\newblock In {\em Proceedings of the {IEEE} conference on computer vision and
  pattern recognition}, pages 883--890.

\bibitem{Zhu_2018}
Yingying Zhu, Jiong Wang, Lingxi Xie, and Liang Zheng.
\newblock Attention-based pyramid aggregation network for visual place
  recognition.
\newblock {\em 2018 ACM Multimedia Conference on Multimedia Conference - MM
  ’18}, 2018.

\bibitem{shi2019deep}
Xiangwei Shi, Seyran Khademi, and Jan van Gemert.
\newblock Deep visual city recognition visualization, 2019.

\bibitem{chopra2005learning}
Sumit Chopra, Raia Hadsell, Yann LeCun, et~al.
\newblock Learning a similarity metric discriminatively, with application to
  face verification.
\newblock In {\em CVPR (1)}, pages 539--546, 2005.

\bibitem{hoffer2015deep}
Elad Hoffer and Nir Ailon.
\newblock Deep metric learning using triplet network.
\newblock In {\em International Workshop on Similarity-Based Pattern
  Recognition}, pages 84--92. Springer, 2015.

\bibitem{nakka2018deep}
Krishna~Kanth Nakka and Mathieu Salzmann.
\newblock Deep attentional structured representation learning for visual
  recognition.
\newblock {\em arXiv preprint arXiv:1805.05389}, 2018.

\bibitem{Lowe2004}
David~G. Lowe.
\newblock Distinctive image features from scale-invariant keypoints.
\newblock {\em International Journal of Computer Vision}, 60(2):91--110, Nov
  2004.

\bibitem{bay2006surf}
Herbert Bay, Tinne Tuytelaars, and Luc Van~Gool.
\newblock Surf: Speeded up robust features.
\newblock In {\em European conference on computer vision}, pages 404--417.
  Springer, 2006.

\bibitem{perronnin2010improving}
Florent Perronnin, Jorge S{\'a}nchez, and Thomas Mensink.
\newblock Improving the fisher kernel for large-scale image classification.
\newblock In {\em European conference on computer vision}, pages 143--156.
  Springer, 2010.

\bibitem{jegou2010aggregating}
Herv{\'e} J{\'e}gou, Matthijs Douze, Cordelia Schmid, and Patrick P{\'e}rez.
\newblock Aggregating local descriptors into a compact image representation.
\newblock In {\em Computer Vision and Pattern Recognition (CVPR), 2010 IEEE
  Conference on}, pages 3304--3311. IEEE, 2010.

\bibitem{majdik_mav_2013}
András~L. Majdik, Yves Albers-Schoenberg, and Davide Scaramuzza.
\newblock Mav urban localization from google street view data.
\newblock In {\em Intelligent Robots and Systems ({IROS}), 2013 {IEEE}/{RSJ}
  International Conference on}, pages 3979--3986. {IEEE}.

\bibitem{chen2017deep}
Zetao Chen, Adam Jacobson, Niko S{\"u}nderhauf, Ben Upcroft, Lingqiao Liu,
  Chunhua Shen, Ian Reid, and Michael Milford.
\newblock Deep learning features at scale for visual place recognition.
\newblock In {\em 2017 IEEE International Conference on Robotics and Automation
  (ICRA)}, pages 3223--3230. IEEE, 2017.

\bibitem{noh2017large}
Hyeonwoo Noh, Andre Araujo, Jack Sim, Tobias Weyand, and Bohyung Han.
\newblock Large-scale image retrieval with attentive deep local features.
\newblock In {\em Proceedings of the IEEE International Conference on Computer
  Vision}, pages 3456--3465, 2017.

\bibitem{sarlin2019coarse}
Paul-Edouard Sarlin, Cesar Cadena, Roland Siegwart, and Marcin Dymczyk.
\newblock From coarse to fine: Robust hierarchical localization at large scale.
\newblock In {\em Proceedings of the IEEE Conference on Computer Vision and
  Pattern Recognition}, pages 12716--12725, 2019.

\bibitem{Babenko_2014}
Artem Babenko, Anton Slesarev, Alexandr Chigorin, and Victor Lempitsky.
\newblock Neural codes for image retrieval.
\newblock {\em Lecture Notes in Computer Science}, page 584–599, 2014.

\bibitem{Yi_2016}
Kwang~Moo Yi, Eduard Trulls, Vincent Lepetit, and Pascal Fua.
\newblock Lift: Learned invariant feature transform.
\newblock {\em Lecture Notes in Computer Science}, page 467–483, 2016.

\bibitem{tolias2015particular}
Giorgos Tolias, Ronan Sicre, and Hervé Jégou.
\newblock Particular object retrieval with integral max-pooling of cnn
  activations, 2015.

\bibitem{gordo2016deep}
Albert Gordo, Jon Almaz{\'a}n, Jerome Revaud, and Diane Larlus.
\newblock Deep image retrieval: Learning global representations for image
  search.
\newblock In {\em European Conference on Computer Vision}, pages 241--257.
  Springer, 2016.

\bibitem{chorowski2015attention}
Jan~K Chorowski, Dzmitry Bahdanau, Dmitriy Serdyuk, Kyunghyun Cho, and Yoshua
  Bengio.
\newblock Attention-based models for speech recognition.
\newblock In {\em Advances in neural information processing systems}, pages
  577--585, 2015.

\bibitem{vaswani2017attention}
Ashish Vaswani, Noam Shazeer, Niki Parmar, Jakob Uszkoreit, Llion Jones,
  Aidan~N Gomez, {\L}ukasz Kaiser, and Illia Polosukhin.
\newblock Attention is all you need.
\newblock In {\em Advances in neural information processing systems}, pages
  5998--6008, 2017.

\bibitem{gu2018attentionaware}
Yinzheng Gu, Chuanpeng Li, and Jinbin Xie.
\newblock Attention-aware generalized mean pooling for image retrieval, 2018.

\bibitem{Long_2018}
Xiang Long, Chuang Gan, Gerard de~Melo, Jiajun Wu, Xiao Liu, and Shilei Wen.
\newblock Attention clusters: Purely attention based local feature integration
  for video classification.
\newblock {\em 2018 IEEE/CVF Conference on Computer Vision and Pattern
  Recognition}, Jun 2018.

\bibitem{seymour2018semanticallyaware}
Zachary Seymour, Karan Sikka, Han-Pang Chiu, Supun Samarasekera, and Rakesh
  Kumar.
\newblock Semantically-aware attentive neural embeddings for image-based visual
  localization, 2018.

\bibitem{song2017deep}
Jifei Song, Qian Yu, Yi-Zhe Song, Tao Xiang, and Timothy~M Hospedales.
\newblock Deep spatial-semantic attention for fine-grained sketch-based image
  retrieval.
\newblock In {\em Proceedings of the IEEE International Conference on Computer
  Vision}, pages 5551--5560, 2017.

\bibitem{xiao2015application}
Tianjun Xiao, Yichong Xu, Kuiyuan Yang, Jiaxing Zhang, Yuxin Peng, and Zheng
  Zhang.
\newblock The application of two-level attention models in deep convolutional
  neural network for fine-grained image classification.
\newblock In {\em Proceedings of the IEEE Conference on Computer Vision and
  Pattern Recognition}, pages 842--850, 2015.

\bibitem{kim2018attention}
Wonsik Kim, Bhavya Goyal, Kunal Chawla, Jungmin Lee, and Keunjoo Kwon.
\newblock Attention-based ensemble for deep metric learning.
\newblock In {\em Proceedings of the European Conference on Computer Vision
  (ECCV)}, pages 736--751, 2018.

\bibitem{Kang_2018}
Guoliang Kang, Liang Zheng, Yan Yan, and Yi~Yang.
\newblock Deep adversarial attention alignment for unsupervised domain
  adaptation: The benefit of target expectation maximization.
\newblock {\em Lecture Notes in Computer Science}, page 420–436, 2018.

\bibitem{wang2019transferable}
Ximei Wang, Liang Li, Weirui Ye, Mingsheng Long, and Jianmin Wang.
\newblock Transferable attention for domain adaptation.
\newblock In {\em AAAI Conference on Artificial Intelligence (AAAI)}, 2019.

\bibitem{long2015learning}
Mingsheng Long, Yue Cao, Jianmin Wang, and Michael~I Jordan.
\newblock Learning transferable features with deep adaptation networks.
\newblock {\em arXiv preprint arXiv:1502.02791}, 2015.

\bibitem{tzeng2014deep}
Eric Tzeng, Judy Hoffman, Ning Zhang, Kate Saenko, and Trevor Darrell.
\newblock Deep domain confusion: Maximizing for domain invariance.
\newblock {\em arXiv preprint arXiv:1412.3474}, 2014.

\bibitem{borgwardt2006integrating}
Karsten~M Borgwardt, Arthur Gretton, Malte~J Rasch, Hans-Peter Kriegel,
  Bernhard Sch{\"o}lkopf, and Alex~J Smola.
\newblock Integrating structured biological data by kernel maximum mean
  discrepancy.
\newblock {\em Bioinformatics}, 22(14):e49--e57, 2006.

\bibitem{gretton2012optimal}
Arthur Gretton, Dino Sejdinovic, Heiko Strathmann, Sivaraman Balakrishnan,
  Massimiliano Pontil, Kenji Fukumizu, and Bharath~K Sriperumbudur.
\newblock Optimal kernel choice for large-scale two-sample tests.
\newblock In {\em Advances in neural information processing systems}, pages
  1205--1213, 2012.

\bibitem{arandjelovic2013all}
Relja Arandjelovic and Andrew Zisserman.
\newblock All about vlad.
\newblock In {\em Proceedings of the IEEE conference on Computer Vision and
  Pattern Recognition}, pages 1578--1585, 2013.

\bibitem{mapilllary}
https://www.mapillary.com.

\bibitem{beeldbank}
https://beeldbank.amsterdam.nl/beeldbank.

\bibitem{deng2009imagenet}
Jia Deng, Wei Dong, Richard Socher, Li-Jia Li, Kai Li, and Li~Fei-Fei.
\newblock Imagenet: A large-scale hierarchical image database.
\newblock In {\em 2009 IEEE conference on computer vision and pattern
  recognition}, pages 248--255. Ieee, 2009.

\bibitem{Baktashmotlagh2019OPENSETDA}
Mahsa Baktashmotlagh, Masoud Faraki, and Tom Drummond.
\newblock Open-set domain adaptation.
\newblock 2019.

\bibitem{Bendale_2016}
Abhijit Bendale and Terrance~E. Boult.
\newblock Towards open set deep networks.
\newblock {\em 2016 IEEE Conference on Computer Vision and Pattern Recognition
  (CVPR)}, Jun 2016.

\bibitem{oza2019deep}
Poojan Oza and Vishal~M. Patel.
\newblock Deep cnn-based multi-task learning for open-set recognition, 2019.

\bibitem{Saito_2018}
Kuniaki Saito, Shohei Yamamoto, Yoshitaka Ushiku, and Tatsuya Harada.
\newblock Open set domain adaptation by backpropagation.
\newblock {\em Lecture Notes in Computer Science}, page 156–171, 2018.

\bibitem{yamaguchi2019adaflow}
Masataka Yamaguchi, Yuma Koizumi, and Noboru Harada.
\newblock Adaflow: Domain-adaptive density estimator with application to
  anomaly detection and unpaired cross-domain translation.
\newblock In {\em ICASSP 2019-2019 IEEE International Conference on Acoustics,
  Speech and Signal Processing (ICASSP)}, pages 3647--3651. IEEE, 2019.

\bibitem{luo2018discriminative}
Lingkun Luo, Liming Chen, Shiqiang Hu, et~al.
\newblock Discriminative label consistent domain adaptation.
\newblock {\em arXiv preprint arXiv:1802.08077}, 2018.

\end{thebibliography}






\end{document}